\pdfoutput=1

\documentclass{article}
\usepackage{colm2024_conference}

\usepackage{url}
\usepackage{booktabs}
\usepackage{hyperref}
\usepackage{graphicx}
\usepackage{microtype}
\usepackage{multicol}
\usepackage{xcolor}   
\usepackage{multirow}
\usepackage{lipsum}
\usepackage{enumitem}
\usepackage{pifont}
\usepackage{colortbl}
\usepackage{tabularx}
\usepackage{caption} 
\usepackage{float}
\usepackage{amsmath,amsfonts,amssymb,bbm}

\newcommand{\tabincell}[2]{\begin{tabular}{@{}#1@{}}#2\end{tabular}}
\newcommand*\samethanks[1][\value{footnote}]{\footnotemark[#1]}
\newcommand{\squad}{\quad \quad }

\title{Stabilizing Reinforcement Learning with LLMs: \\ Formulation and Practices}

\author{
\textbf{Chujie Zheng\thanks{Corresponding authors.} \squad Kai Dang \squad Bowen Yu\samethanks{} \squad Mingze Li \squad Huiqiang Jiang \\ Junrong Lin \squad Yuqiong Liu \squad Hao Lin \squad Chencan Wu \squad Feng Hu \\ An Yang \squad Jingren Zhou \squad Junyang Lin} \\
\vspace{1.5mm}
Qwen Team, Alibaba Inc.
}

\begin{document}
\maketitle

\begin{abstract}

This paper proposes a novel formulation for reinforcement learning (RL) with large language models, explaining why and under what conditions the true sequence-level reward can be optimized via a surrogate token-level objective in policy gradient methods such as REINFORCE.
Specifically, through a first-order approximation, we show that this surrogate becomes increasingly valid only when both the training–inference discrepancy and policy staleness are minimized.
This insight provides a principled explanation for the crucial role of several widely adopted techniques in stabilizing RL training, including importance sampling correction, clipping, and particularly Routing Replay for Mixture-of-Experts (MoE) models.
Through extensive experiments with a 30B MoE model totaling hundreds of thousands of GPU hours, we show that for on-policy training, the basic policy gradient algorithm with importance sampling correction achieves the highest training stability.
When off-policy updates are introduced to accelerate convergence, combining clipping and Routing Replay becomes essential to mitigate the instability caused by policy staleness.
Notably, once training is stabilized, prolonged optimization consistently yields comparable final performance regardless of cold-start initialization.
We hope that the shared insights and the developed recipes for stable RL training will facilitate future research.

\end{abstract}

\section{Introduction}
\label{sec:intro}

Reinforcement learning (RL) has become a key technical paradigm for enhancing large language models' (LLMs) ability to tackle complex problem-solving tasks \citep{o1,dpsk-r1,qwen3}, while a stable training process\footnote{
By \textit{stable training}, we refer to a training process in which model performance steadily improves over training steps—reflected in both the training reward and benchmark scores—and, crucially, the model’s internal state evolves smoothly and without abrupt shifts.
The model state can be monitored via a set of diagnostic metrics, including the training-inference KL divergence and entropy reported in our later experiments (\S~\ref{sec:exp}).
Stable training implies that the model can consistently improve in a healthy manner throughout extended or multi-stage training, with a lower risk of unexpected behaviors.
} is crucial for successfully scaling RL.
Due to the contextual nature of language, RL with LLMs usually employs sequence-level rewards, i.e., a scalar score assigned based on the complete model response.
However, mainstream RL algorithms, such as REINFORCE and GRPO, typically employ token-level optimization objectives.
This mismatch between the reward (assigned at the sequence level) and the optimization unit (typically at the token level) raises concerns about the soundness and training stability of such approaches, while some studies have proposed directly adopting sequence-level optimization objectives \citep{gspo,liu-li-2025-rl-collapse}.
In particular, token-level optimization objectives also pose unique challenges for RL training with Mixture-of-Experts (MoE) models.
For instance, the dynamic expert routing mechanism can invalidate the token-level importance sampling ratios in MoE models \citep{gspo}.
However, it remains unclear whether optimizing sequence-level rewards using token-level objectives is justified, and if so, to what extent (or under what conditions) such an approach is valid.

In this paper, we propose a novel formulation for RL with LLMs.
The key insight is that, to optimize the expected sequence-level reward, we can employ a surrogate token-level objective as its first-order approximation.
Specifically, this approximation is likely to hold only when both (1) the numerical discrepancy between the training and inference engines (i.e., the training–inference discrepancy) and (2) the discrepancy between the rollout policy that samples responses and the target policy to be optimized (i.e., policy staleness) are minimized.
This insight provides a principled explanation of how several techniques for stabilizing RL training work.
For example, 
(1) the importance sampling weight is an inherent component of the surrogate token-level objective under the first-order approximation;
(2) the clipping mechanism can restrain policy staleness by preventing aggressive policy updates;
(3) for MoE models, the Routing Replay approach \citep{gspo,r3}, which fixes the routed experts during policy optimization, can reduce both the training–inference discrepancy and policy staleness.

To empirically validate our insight and investigate practical recipes for stable RL training, we conduct extensive experiments with a 30B MoE model, amounting to hundreds of thousands of GPU hours.
Our main conclusions include:
(1) For on-policy training\footnote{
In this paper, we use the term \textit{on-policy} to indicate that the rollout policy that samples responses is identical to the target policy to be optimized using these responses (omitting the training–inference discrepancy), while \textit{off-policy} indicates that the two policies are different.
}, the basic policy gradient algorithm with importance sampling correction yields the highest training stability;
(2) When off-policy updates are introduced to accelerate convergence, i.e., a large batch of responses is split into mini-batches for multiple gradient updates, combining clipping and Routing Replay becomes necessary to mitigate instability caused by policy staleness;
(3) Once training is stabilized, models with different cold-start initializations consistently achieve comparable final performance.
This motivates future work to focus more on RL itself rather than overly on the specifics of cold-start initialization, as differences arising from the latter are expected to vanish given prolonged RL training.

In summary, this paper makes contributions along two axes:
\begin{itemize}[leftmargin=10mm, itemsep=2mm]

\item Theoretically, we propose a novel formulation for reinforcement learning with LLMs, revealing the conditions under which optimizing sequence-level rewards via token-level objectives is justified.
Specifically, the validity of the underlying first-order approximation hinges on jointly minimizing the training–inference discrepancy and policy staleness.

\item Empirically, through extensive experiments with MoE models spanning hundreds of thousands of GPU hours, we demonstrate that several techniques that preserve the validity of the first-order approximation consistently exhibit practical efficacy in stabilizing RL training, particularly the Routing Replay approach tailored for MoE models.
We hope that the developed recipes for stable RL training will facilitate future research.

\end{itemize}

\section{Formulation for Reinforcement Learning with LLMs}
\label{sec:formulation}

\subsection{Notation}

We define an autoregressive LLM parameterized by \( \theta \) as a policy \( \pi_\theta \).
We use \(x\) to denote an input prompt and \( \mathcal{D} \) as the prompt set.
Under the policy \( \pi_\theta \), the likelihood of a response \(y\) to a prompt \(x\) is denoted as \( \pi_\theta (y | x)=\prod_{t=1}^{|y|} \pi_\theta (y_t | x, y_{<t} ) \) where \( |y| \) is the number of tokens in \(y\).
Given the contextual nature of language, we focus on the sequence-level reward setting, where a whole response \(y\) is assigned a single scalar reward \( R(x,y) \).
We do not consider the value-based setting (e.g., PPO, \citealt{ppo}), where policy optimization is steered by a value model that assigns scalar scores to each token in a response \(y\).
This is because we found it inherently difficult (if not impossible) to devise general and scalable approaches to obtaining reliable value models.

\subsection{Expected Sequence-level Reward is Hard to Directly Optimize}

Our formulation starts from the true sequence-level reward that we aim to maximize:
\begin{align*}
\mathcal{J}^\text{seq}(\theta)
=
\mathbb{E}_{ x \sim \mathcal{D}, \, y \sim \pi_\theta ( \cdot | x ) }  \left[ R(x, y) \right],
\end{align*}
where \( \pi_\theta \) is the target policy to be optimized.
Since the responses are typically not sampled in the training engine (e.g., Megatron and FSDP) but instead in the inference engine (e.g., SGLang and vLLM), we adopt the importance sampling (IS) trick to do a simple transformation:
\begin{align}
\label{equ:seq_obj}
\mathcal{J}^\text{seq}(\theta)
=
\mathbb{E}_{ x \sim \mathcal{D}, \, y \sim \pi_\theta ( \cdot | x ) }  \left[ R(x, y) \right] 
= 
\mathbb{E}_{ x \sim \mathcal{D}, \, y \sim \mu_{ \theta_\text{old} } ( \cdot | x ) } \left[ 
\underbrace{
    \frac{ \pi_\theta ( y | x ) }{ \mu_{ \theta_\text{old} } ( y | x ) }
}_\text{sequence-level IS weight}
R(x, y) \right],
\end{align}
where \( \mu_{ \theta_\text{old} } \) denotes the rollout policy that samples responses.
Note that we use the notation \( \mu \) to distinguish the policy in the inference engine from the policy (notated as \( \pi \)) in the training engine, as there typically exists a numerical discrepancy between training and inference engines \citep{yao2025offpolicy}.
The sequence-level objective in Equation~\eqref{equ:seq_obj} has the following gradient:
\begin{align}
\nabla_\theta \, \mathcal{J}^\text{seq}(\theta)
= &\ 
\mathbb{E}_{ x \sim \mathcal{D}, \, y \sim \mu_{ \theta_\text{old} } ( \cdot | x ) } \left[ \frac{ \pi_\theta ( y | x ) }{ \mu_{ \theta_\text{old} } ( y | x ) } \, R(x, y) \, \nabla_\theta \log \pi_\theta ( y | x ) \right]  \notag \\
\label{equ:seq_grad}
= &\ 
\mathbb{E}_{ x \sim \mathcal{D}, \, y \sim \mu_{ \theta_\text{old} } ( \cdot | x ) } \left[ \frac{ \pi_\theta ( y | x ) }{ \mu_{ \theta_\text{old} } ( y | x ) } \, R(x, y) \sum_{t=1}^{|y|} \nabla_\theta \log \pi_\theta ( y_t | x, y_{<t} ) \right].
\end{align}
However, this gradient is usually intractable to utilize due to the large numerical range and high variance of sequence likelihood (i.e., \( \pi_\theta ( y | x ) \) and \( \mu_{ \theta_\text{old} } ( y | x ) \)), making it difficult to directly optimize the sequence-level objective in Equation~\eqref{equ:seq_obj}.

\subsection{Token-level Objective as a First-order Approximation to Sequence-level Objective}

The critical step in our formulation is to consider the following surrogate token-level objective:
\begin{align}
\label{equ:token_obj}
\mathcal{J}^\text{token}(\theta)
=
\mathbb{E}_{ x \sim \mathcal{D}, \, y \sim \mu_{ \theta_\text{old} } ( \cdot | x ) } \left[ \sum_{t=1}^{|y|} 
\underbrace{
    \frac{ \pi_\theta ( y_t | x, y_{<t} ) }{ \mu_{ \theta_\text{old} } ( y_t | x, y_{<t} ) }
}_\text{token-level IS weight}
R(x, y)
\right],
\end{align}
with the following gradient:
\begin{align}
\label{equ:token_grad}
\nabla_\theta \, \mathcal{J}^\text{token}(\theta)
=
\mathbb{E}_{ x \sim \mathcal{D}, \, y \sim \mu_{ \theta_\text{old} } ( \cdot | x ) }
\left[ \sum_{t=1}^{|y|} \frac{ \pi_\theta ( y_t | x, y_{<t} ) }{ \mu_{ \theta_\text{old} } ( y_t | x, y_{<t} ) } \, R(x, y) \, \nabla_\theta \log \pi_\theta ( y_t | x, y_{<t} ) \right].
\end{align}
which is actually the basic policy gradient algorithm (i.e., REINFORCE) equipped with a token-level IS weight.
The core insight here is that \textbf{we can view the token-level optimization objective in Equation~\eqref{equ:token_obj} as a first-order approximation to the sequence-level objective in Equation~\eqref{equ:seq_obj} that we truly aim to optimize}.
To be specific, suppose \( \pi_\theta \) and \( \mu_{ \theta_\text{old} } \) are slightly different, let \( \frac{ \pi_\theta ( y_t | x, y_{<t} ) }{ \mu_{ \theta_\text{old} } ( y_t | x, y_{<t} ) } = 1 + \delta_t \) where \( \delta_t \) is a small quantity.
We have the following approximation:
\begin{align*}
\frac{ \pi_\theta ( y | x ) }{ \mu_{ \theta_\text{old} } ( y | x ) }
=
\prod_{t=1}^{|y|} \left( 1 + \delta_t \right)
\approx
1 + \sum_{t=1}^{|y|} \delta_t + O \left( \delta^2 \right)
\approx
1 + \sum_{t=1}^{|y|} \delta_t,
\end{align*}
where the rightmost derivation neglects second-order and higher-order small terms like \(\delta_i \delta_j\).
So we have:
\begin{align*}
\nabla_\theta \, \mathcal{J}^\text{seq}(\theta)
= &\
\mathbb{E}_{ x \sim \mathcal{D}, \, y \sim \mu_{ \theta_\text{old} } ( \cdot | x ) } \left[ 
    R(x, y) \, \nabla_\theta \left(
    \frac{ \pi_\theta ( y | x ) }{ \mu_{ \theta_\text{old} } ( y | x ) }
    \right)
\right] \\
\approx &\
\mathbb{E}_{ x \sim \mathcal{D}, \, y \sim \mu_{ \theta_\text{old} } ( \cdot | x ) } \left[ 
    R(x, y) \, \nabla_\theta \left(
    1 + \sum_{t=1}^{|y|} \delta_t
    \right)
\right] \\
= &\
\mathbb{E}_{ x \sim \mathcal{D}, \, y \sim \mu_{ \theta_\text{old} } ( \cdot | x ) } \left[ 
    R(x, y) \, \nabla_\theta \left(
    \sum_{t=1}^{|y|} \frac{ \pi_\theta ( y_t | x, y_{<t} ) }{ \mu_{ \theta_\text{old} } ( y_t | x, y_{<t} ) }
    \right)
\right] \\
= &\
\nabla_\theta \, \mathcal{J}^\text{token}(\theta).
\end{align*}
This is why we say that Equation~\eqref{equ:token_obj} is a \textit{first-order} approximation to Equation~\eqref{equ:seq_obj}.
Therefore, \textbf{when \( \pi_\theta \) is close to \( \mu_{ \theta_\text{old} } \), we can improve the sequence-level objective in Equation~\eqref{equ:seq_obj} by updating the model parameters \( \theta \) with the gradient in Equation~\eqref{equ:token_grad}}.

\subsection{Conditions for First-order Approximation to Hold}

For the first-order approximation to hold, we require that the target policy \( \pi_\theta \) and the rollout policy \( \mu_{ \theta_\text{old} } \) are close, which, however, is less intuitive.
To be clear, given \(x\) and for each token \( y_t \), we rewrite its IS weight as:
\begin{align}
\label{equ:is_decomp}
\frac{ \pi_\theta ( y_t | x, y_{<t} ) }{ \mu_{ \theta_\text{old} } ( y_t | x, y_{<t} ) }
=
\underbrace{
    \frac{ \pi_{ \theta_\text{old} } ( y_t | x, y_{<t} ) }{ \mu_{ \theta_\text{old} } ( y_t | x, y_{<t} ) } 
}_\text{training–inference discrepancy}
\times 
\underbrace{
\frac{ \pi_\theta ( y_t | x, y_{<t} ) }{ \pi_{ \theta_\text{old} } ( y_t | x, y_{<t} ) }
}_\text{policy staleness},
\end{align}
where \( \pi_{ \theta_\text{old} } \) denotes the rollout policy computed by the training engine, differing from the one \( \mu_{ \theta_\text{old} } \) in the inference engine.
Therefore, from the decomposition in Equation~\eqref{equ:is_decomp}, the gap between \( \pi_\theta \) and \( \mu_{ \theta_\text{old} } \) comes from two aspects: the \textit{training–inference discrepancy} and \textit{policy staleness}.

\begin{itemize}[leftmargin=10mm, itemsep=2mm]

\item Regarding the \textbf{training–inference discrepancy}—i.e., the numerical differences between training and inference engines—the causes are usually complex and heavily tied to the underlying infrastructure. 
For example, training and inference engines typically employ different computational kernels for peak performance, which would yield inconsistent outputs given the same model input.
Even within a single engine, particularly the inference side, batch-invariant kernels \citep{he2025nondeterminism} are often disabled for maximizing throughput, so the same model input can still receive variant outputs.
In the case of MoE models, the training–inference discrepancy is further amplified by inconsistent expert routing, which we will discuss detailedly in \S~\ref{sec:moe}.

\item Regarding \textbf{policy staleness}—i.e., the discrepancy between the rollout policy that samples responses and the target policy to be optimized—it usually arises from the trade-offs made to improve training efficiency and computational utilization.
Since the rollout stage in RL is typically bounded in time by the generation length, to accelerate convergence through increased computational resources, we often split a large batch of sampled responses into mini-batches for multiple gradient updates.
Consequently, mini-batches consumed later may exhibit greater policy staleness.
In asynchronous RL frameworks, a single response can be generated sequentially by multiple model versions, which also introduces policy staleness.

\end{itemize}

Therefore, to ensure the validity of the first-order approximation that underlies the surrogate token-level objective in Equation~\eqref{equ:token_obj}, we should, in principle, narrow the gap between \( \pi_\theta \) and \( \mu_{ \theta_\text{old} } \) from two directions: {reducing the numerical discrepancy between training and inference engines}, and {controlling policy staleness within a moderate range}.

\section{Challenge for Mixture of Experts, and Routing Replay}
\label{sec:moe}

\subsection{Expert Routing Hinders First-order Approximation to Hold}

When it comes to Mixture-of-Experts (MoE) models \citep{dpsk-r1,qwen3}, the conditions for the first-order approximation to hold become less straightforward.
Specifically, during the forward pass of generating each token, MoE models dynamically select and activate only a small subset of expert parameters via the expert routing mechanism.
Incorporating expert routing into Equation~\eqref{equ:is_decomp}, we can write the token-level IS weight for an MoE model as:
\begin{align}
\label{equ:is_decomp_moe}
\frac{ \pi_\theta ( y_t | x, y_{<t} ) }{ \mu_{ \theta_\text{old} } ( y_t | x, y_{<t} ) }
=
\frac{ \pi_\theta ( y_t | x, y_{<t}, \textcolor{red}{ e^{ \pi }_t } ) }{ \mu_{ \theta_\text{old} } ( y_t | x, y_{<t}, \textcolor{red}{ e^{ \mu }_{ \text{old}, \, t } } ) }
=
\underbrace{
    \frac{ \pi_{ \theta_\text{old} } ( y_t | x, y_{<t}, \textcolor{red}{ e^{ \pi }_{ \text{old}, \, t } } ) }{ \mu_{ \theta_\text{old} } ( y_t | x, y_{<t}, \textcolor{red}{ e^{ \mu }_{ \text{old}, \, t } } ) } 
}_\text{training–inference discrepancy}
\times 
\underbrace{
\frac{ \pi_\theta ( y_t | x, y_{<t}, \textcolor{red}{ e^{ \pi }_t } ) }{ \pi_{ \theta_\text{old} } ( y_t | x, y_{<t}, \textcolor{red}{ e^{ \pi }_{ \text{old}, \, t } } ) }
}_\text{policy staleness},
\end{align}
where \( e^{ \pi } \) and \( e^{ \mu } \) denote the routed experts in the training and inference engines, respectively, and the subscript ``old'' corresponds to the rollout policy.

At this point, the challenge of reinforcement learning with MoE models becomes clear: {expert routing is entangled with the training–inference discrepancy and policy staleness, increasing the likelihood that the first-order approximation underlying the surrogate token-level optimization objective in Equation~\eqref{equ:token_obj} breaks down}.
More specifically, the training–inference discrepancy can cause inconsistent routed experts in the training and inference engines (i.e., \( e^{ \pi }_{ \text{old}, \, t } \) versus \( e^{ \mu }_{ \text{old}, \, t } \)) given the same model parameters and input.
This divergence in expert routing, in turn, further amplifies the discrepancy in final outputs.
Furthermore, policy staleness manifests not only in changes in the model parameters (i.e., \( \theta \) versus \( \theta_\text{old} \)) but also in shifts of routed experts (i.e., \( e^{ \pi }_t \) versus \( e^{ \pi }_{ \text{old}, \, t } \)), which can heavily alter the resulting policy defined by activated parameters.

\subsection{Routing Replay Restores First-order Approximation, Yet May Introduce Bias}
\label{subsec:routing_replay}

Identifying that expert routing undermines the validity of the first-order approximation in MoE models, we can eliminate this impact through the Routing Replay \citep{gspo} approach.
The core idea of Routing Replay is to stabilize RL training of MoE models by fixing the routed experts during policy optimization, thereby enabling the model to be optimized like a dense one.
Upon Equation~\eqref{equ:is_decomp_moe}, we formalize the following two concrete implementations of Routing Replay, namely \textit{Vanilla Routing Replay} and \textit{Rollout Routing Replay}:
\begin{itemize}[leftmargin=10mm, itemsep=2mm]

\item \textbf{Vanilla Routing Replay (R2)} \citep{gspo} focuses on mitigating the impact of expert routing on policy staleness by replaying, during gradient updates, the routed experts determined by the rollout policy in the training engine (i.e., \( e^{ \pi }_{ \text{old}, \, t } \)):
\begin{align*}
\frac{ \pi^\text{R2}_\theta ( y_t | x, y_{<t} ) }{ \mu_{ \theta_\text{old} } ( y_t | x, y_{<t} ) }
=
\frac{ \pi_\theta ( y_t | x, y_{<t}, \textcolor{blue}{ e^{ \pi }_{ \text{old}, \, t } } ) }{ \mu_{ \theta_\text{old} } ( y_t | x, y_{<t}, e^{ \mu }_{ \text{old}, \, t } ) }
=
\frac{ \pi_{ \theta_\text{old} } ( y_t | x, y_{<t}, e^{ \pi }_{ \text{old}, \, t } ) }{ \mu_{ \theta_\text{old} } ( y_t | x, y_{<t}, e^{ \mu }_{ \text{old}, \, t } ) } 
\times 
\underbrace{
    \frac{ \pi_\theta ( y_t | x, y_{<t}, \textcolor{blue}{ e^{ \pi }_{ \text{old}, \, t } } ) }{ \pi_{ \theta_\text{old} } ( y_t | x, y_{<t}, e^{ \pi }_{ \text{old}, \, t } ) }
}_\text{policy staleness ↓}.
\end{align*}

\item \textbf{Rollout Routing Replay (R3)} \citep{r3} aims to reduce the impact of expert routing on the training–inference discrepancy by uniformly replaying, within the training engine, the routed experts determined by the rollout policy in the inference engine (i.e., \( e^{ \mu }_{ \text{old}, \, t } \)), which also simultaneously mitigates the impact of expert routing on policy staleness:
\begin{align*}
\frac{ \pi^\text{R3}_\theta ( y_t | x, y_{<t} ) }{ \mu_{ \theta_\text{old} } ( y_t | x, y_{<t} ) }
=
\frac{ \pi_\theta ( y_t | x, y_{<t}, \textcolor{blue}{ e^{ \mu }_{ \text{old}, \, t } } ) }{ \mu_{ \theta_\text{old} } ( y_t | x, y_{<t}, e^{ \mu }_{ \text{old}, \, t } ) }
=
\underbrace{
    \frac{ \pi_{ \theta_\text{old} } ( y_t | x, y_{<t}, \textcolor{blue}{ e^{ \mu }_{ \text{old}, \, t } } ) }{ \mu_{ \theta_\text{old} } ( y_t | x, y_{<t}, e^{ \mu }_{ \text{old}, \, t } ) } 
}_\text{training–inference discrepancy ↓}
\times 
\underbrace{
    \frac{ \pi_\theta ( y_t | x, y_{<t}, \textcolor{blue}{ e^{ \mu }_{ \text{old}, \, t } } ) }{ \pi_{ \theta_\text{old} } ( y_t | x, y_{<t}, \textcolor{blue}{ e^{ \mu }_{ \text{old}, \, t } } ) }
}_\text{policy staleness ↓}.
\end{align*}

\end{itemize}

Therefore, {Routing Replay intuitively restores the validity of the first-order approximation in MoE models by reducing the training–inference discrepancy (in R3) and alleviating policy staleness (in R2 and R3)}.
However, we point out that {it also implicitly biases the target policy}, as suggested by the notations \( \pi^\text{R2}_\theta \) and \( \pi^\text{R3}_\theta \).
Specifically, the original target policy we aime to optimize in Equation~\eqref{equ:token_obj} is \( \pi_\theta \), where the likelihood of each token \( y_t \) should be governed by the \textit{naturally}-routed experts \( e^{\pi}_t \).
However, Routing Replay constrains the routed experts to be \( e^{ \pi }_{ \text{old}, \, t } \) or \( e^{ \mu }_{ \text{old}, \, t } \), leading to another target policy \( \pi^\text{R2}_\theta \) or \( \pi^\text{R3}_\theta \) that deviates from the original \( \pi_\theta \) defined by \( e^{\pi}_t \).
In particular, when we split a large batch into mini-batches for multiple gradient updates, R2 and R3 can possess different degrees of bias, as shown in Table~\ref{tab:compare_r2_r3}.
The key difference is that R2 does not alter the original target policy in the first mini-batch, which we conjecture may lead to R2 and R3 exhibiting different performance, especially when the ratio between batch size and mini-batch size (i.e., the degree of off-policiness) is varied.

\begin{table}[htbp]
    \centering
    \caption{Comparison between R2 and R3 in how they alter the original target policy \( \pi_\theta \).}
    \begin{tabular}{ccc}
        \toprule
    & \tabincell{c}{First mini-batch} & \tabincell{c}{Non-first mini-batch} \\
        \midrule
    \tabincell{c}{R2 \\ (replaying \( e^{ \pi }_{ \text{old}, \, t } \))} &  \tabincell{c}{ \textcolor{blue}{ \( e^{ \pi }_{ \text{old}, \, t } = e^{\pi}_t \), } \\ \textcolor{blue}{ target policy is not altered }} & \tabincell{c}{ \( e^{ \pi }_{ \text{old}, \, t } \ne e^{\pi}_t \), \\ target policy is altered} \\
    \midrule[0mm]
    \tabincell{c}{R3 \\ (replaying \( e^{ \mu }_{ \text{old}, \, t } \))} & \tabincell{c}{ \( e^{ \mu }_{ \text{old}, \, t } \ne e^{\pi}_t \), \\ target policy is altered}  & \tabincell{c}{ \( e^{ \mu }_{ \text{old}, \, t } \ne e^{\pi}_t \), \\ target policy is altered} \\
    \bottomrule
    \end{tabular}
    \label{tab:compare_r2_r3}
\end{table}

Nevertheless, it is difficult to definitively assess whether the advantages or disadvantages of Routing Replay outweigh each other.
Altering routed experts, while introducing bias into the optimization objective, also makes the first-order approximation—on which the altered token-level objective using \( \pi^\text{R2}_\theta \) or \( \pi^\text{R3}_\theta \) as the target policy relies—more likely to hold.
We need further experiments to validate the practical utility of Routing Replay.

\section{Empirical Analyses}
\label{sec:exp}

\subsection{MiniRL: A Minimalist Baseline Algorithm}

In our experiments, we employ two minimal modifications to the REINFORCE optimization objective in Equation~\eqref{equ:token_obj} as a minimalist baseline algorithm.
First, we apply group-normalization \citep{grpo} to the raw rewards as the advantage estimate for each response \(y\): \( \widehat{A}(x, y) = R(x, y) - \mathbb{E}_{ y' \sim \mu_{ \theta_\text{old} } ( \cdot | x ) } \left[ R(x, y') \right] \), which also lowers the variance of the raw rewards.
Second, we adopt the clipping mechanism in PPO \citep{ppo} that prevents aggressive policy updates by stopping gradients for certain tokens, which can hopefully restrain policy staleness.
We follow the decoupled PPO approach \citep{hilton2022batch} and use \( \pi_{ \theta_\text{old} } \) as the proximal policy to decide whether to clip the token \( y_t \) based on the ratio of \( \pi_\theta ( y_t | x, y_{<t} ) \) and \( \pi_{ \theta_\text{old} } ( y_t | x, y_{<t} ) \)\footnote{
While there are alternative clipping strategies, such as clipping a whole response based on the ratio of sequence likelihood \citep{gspo}, we found that the current clipping strategy has worked decently.
Therefore, we leave the study of clipping or masking strategies for future work.
Similarly, exploring better advantage estimates \( \widehat{A}(x, y) \) may also be helpful, but falls outside the scope of this work.
}.
The obtained minimalist baseline algorithm, which we call \textbf{MiniRL}, is as follows:
\begin{align}
\label{equ:minirl}
\mathcal{J}_\text{MiniRL}(\theta)
=&\
\mathbb{E}_{ x \sim \mathcal{D}, \, y \sim \mu_{ \theta_\text{old} } ( \cdot | x ) }
\left[ \sum_{t=1}^{|y|} 
M_t \,
\mathrm{sg} \left[ \frac{ \pi_\theta ( y_t | x, y_{<t} ) }{ \mu_{ \theta_\text{old} } ( y_t | x, y_{<t} ) } \right] 
\widehat{A}(x, y) \, 
\log \pi_\theta ( y_t | x, y_{<t} ) \right], \\
\notag
M_t = &\
\begin{cases}
0 & \text{if } \widehat{A}(x, y) > 0 \text{ and } r_t > 1 + \varepsilon_\text{high},  \\
0 & \text{if } \widehat{A}(x, y) < 0 \text{ and } r_t < 1 - \varepsilon_\text{low}, \\
1  & \text{otherwise},
\end{cases} \quad \quad
r_t = \frac{ \pi_\theta ( y_t | x, y_{<t} ) }{ \pi_{ \theta_\text{old} } ( y_t | x, y_{<t} ) },
\end{align}
where $\mathrm{sg}$ denotes the operation of stopping gradient.
It is noteworthy that MiniRL is adopted as the baseline algorithm to maintain consistency—as closely as possible—(in gradient) with the surrogate token-level objective in Equation~\ref{equ:token_obj}, which has been justified by our formulation in \S~\ref{sec:formulation}.
In Appendix~\ref{app:algo}, we will provide a comparison of MiniRL against other algorithms such as GRPO \citep{grpo} and CISPO \citep{minimax-m1}.
All our experiments will be implemented based on MiniRL.

\subsection{Experimental Setup}

We conduct experiments on the mathematical reasoning task, where the model response is compared with the ground truth answer and then assigned a binary reward (i.e., \( R(x, y) \in \{0, 1\} \)).
We curate 4,096 math problems with verified answers as the prompt set for RL training.
We report the average accuracy over 32 sampled responses on the HMMT25, AIME25, and AIME24 benchmarks, each consisting of 30 competition-level math problems (90 in total).

We experiment with a cold-start model fine-tuned from Qwen3-30B-A3B-Base.
We adopt the setting of FP8 inference and BF16 training, providing a stress test for algorithmic correctness where the inference precision is lower than the training and the training–inference discrepancy is large.
Besides the training reward, we also report the dynamics of two metrics:
(1) the token-level entropy of the target policy, approximated by: 
\begin{align*}
\mathbb{H} \left[ \pi_\theta \right] 
\approx  
\mathbb{E}_{ x \sim \mathcal{D}, \, y_{<t} \sim \mu_{ \theta_\text{old} (\cdot | x) } } \left[ - \sum_{w \in \mathcal{V} } \pi_\theta ( w | x,\, y_{<t}) \log \pi_\theta ( w | x,\, y_{<t})
\right],
\end{align*}
where \( \mathcal{V} \) denotes the vocabulary,
and (2) the KL divergence between the rollout policies in the inference and training engines, calculated as: 
\begin{align*}
\mathbb{D}_\text{KL} 
\left[ \mu_{ \theta_\text{old} } \| \pi_{ \theta_\text{old} } \right] 
=
\mathbb{E}_{ x \sim \mathcal{D}, \, y_t \sim \mu_{ \theta_\text{old} (\cdot | x,\, y_{<t}) } } \left[ \log 
\frac{ \mu_{ \theta_\text{old} } ( y_t | x,\, y_{<t} ) }{ \pi_{ \theta_\text{old} } ( y_t | x,\, y_{<t} ) } 
\right].
\end{align*}
We report the latter metric because recent work \citep{yao2025offpolicy,liu-li-2025-rl-collapse} has revealed that the instability or collapse in RL training is often accompanied by a sharp increase in the training-inference discrepancy.

To conduct controlled experiments, we employ the standard synchronous RL framework.
In each global step, we first sample a batch of \(B\) prompts and sample \(G\) responses for each prompt using the rollout policy in the inference engine.
Then, we split the responses into \(N\) mini-batches and apply \(N\) gradient updates in the training engine.
The finally updated policy in this global step is used as the new rollout policy in the next global step.
Across all experimental runs, we use the same mini-batch size of 1,024 responses (\( B=64 \) and \( G=16 \)) for each gradient update.

For other hyperparameters, we set the maximum generation length to 32,768, and set \( \varepsilon_\text{high} \) to 0.27 and \( \varepsilon_\text{low} \) to 0.2 in MiniRL.
We additionally apply the Truncated Importance Sampling (TIS) trick \citep{yao2025offpolicy} to the token-level IS weight in MiniRL, with the truncation threshold set to 5.
Our experiments total hundreds of thousands of GPU hours, and the consumed compute can be estimated as \( 5 \sim 6 \) GPU hours per gradient step.

\subsection{Results of On-policy Training}
\label{subsec:on-policy}

We first verify, under on-policy training where the global batch size equals the mini-batch size, whether the validity of the first-order approximation underlying the token-level optimization objective is correlated with training stability.
Under this \textit{on-policy} setting where  \(\theta = \theta_\text{old}\), MiniRL degenerates to the following basic policy gradient algorithm:
\begin{align*}
\label{equ:minirl_onpolicy}
\mathcal{J}_\text{MiniRL}(\theta)
=&\
\mathbb{E}_{ x \sim \mathcal{D}, \, y \sim \mu_{ \theta_\text{old} } ( \cdot | x ) }
\left[ \sum_{t=1}^{|y|} 
\frac{ \pi_{ \theta_\text{old} } ( y_t | x, y_{<t} ) }{ \mu_{ \theta_\text{old} } ( y_t | x, y_{<t} ) } \,
\widehat{A}(x, y) \, 
\log \pi_\theta ( y_t | x, y_{<t} ) \right],
\end{align*}
so the IS weight here serves only as a correction for the training–inference discrepancy.
We notice that existing RL algorithms, such as GRPO and CISPO, often employ length normalization in their optimization objectives, and their original objectives do not consider IS correction for the training–inference discrepancy.
We thus include the following two ablated variants of MiniRL in our experiments:
\begin{align*}
\mathcal{J}^\text{w\_length-norm}_\text{MiniRL}(\theta)
=
\mathbb{E}_{ x \sim \mathcal{D}, \, y \sim \mu_{ \theta_\text{old} } ( \cdot | x ) } 
\left[ \textcolor{blue}{ \frac{1}{|y|} } \sum_{t=1}^{|y|} 
\frac{ \pi_{ \theta_\text{old} } ( y_t | x, y_{<t} ) }{ \mu_{ \theta_\text{old} } ( y_t | x, y_{<t} ) } \,
\widehat{A}(x, y) \, \log \pi_\theta ( y_t | x, y_{<t} ) \right],
\end{align*}
which additionally employs length normalization, and
\begin{align*}
\mathcal{J}^\text{wo\_train-infer-is}_\text{MiniRL}(\theta)
=
\mathbb{E}_{ x \sim \mathcal{D}, \, y \sim \mu_{ \theta_\text{old} } ( \cdot | x ) } 
\left[ \sum_{t=1}^{|y|} 
\widehat{A}(x, y) \, \log \pi_\theta ( y_t | x, y_{<t} ) \right],
\end{align*}
which omits the IS correction for the training–inference discrepancy.
Note that the two variants have \textit{no longer satisfied the aforementioned first-order approximation}, as their gradients are neither equal to nor linearly correlated with the gradient of the true sequence-level objective in Equation~\eqref{equ:seq_obj} (ignoring the reward normalization).
We also equip MiniRL and the two variants with R3 (R2 is inapplicable here, see Table~\ref{tab:compare_r2_r3}) for comparison.

\begin{figure*}[htbp]
    \centering
    \includegraphics[width=\linewidth]{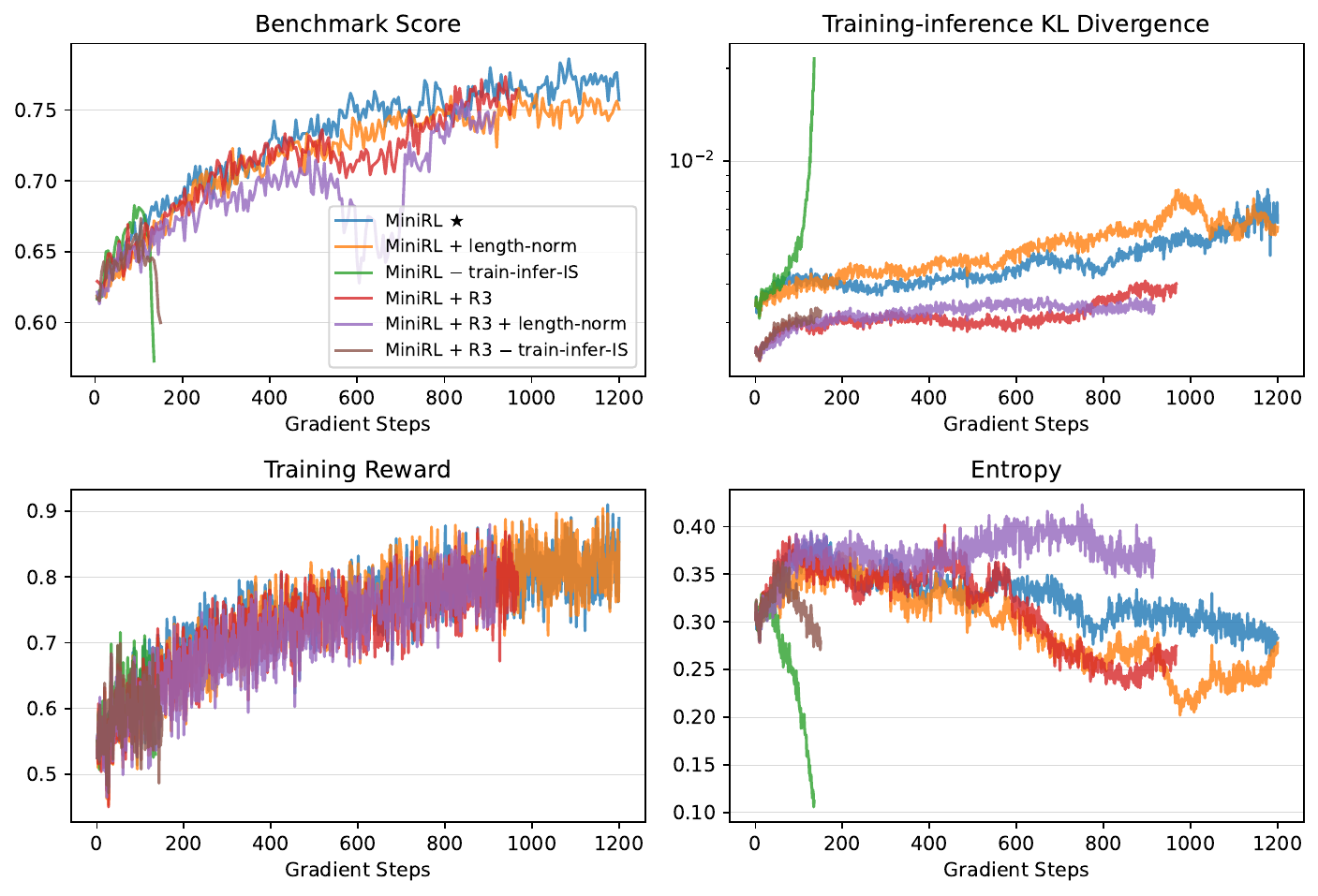}
    \caption{Results of on-policy training with \( \text{gbs (global batch size)}=\text{mbs (mini-batch size)}=1,024 \).}
    \label{fig:gbs=mbs}
\end{figure*}

From Figure~\ref{fig:gbs=mbs}, we draw the following observations and conclusions:
\begin{itemize}[leftmargin=10mm, itemsep=2mm]

\item MiniRL, i.e., the basic policy gradient algorithm with IS correction, achieves the best performance and training stability.

\item Adding length normalization leads to suboptimal performance\footnote{
Our conclusion regarding length normalization aligns with that of \citet{dr.grpo}, albeit with a markedly different motivation.
}, although training remains stable.
This is as expected, since length normalization invalidates the first-order approximation to the true expected sequence-level reward, resulting in a biased token-level optimization objective.

\item Removing the training–inference IS correction causes rapid training collapse and a sharp drop in entropy.
This confirms that the IS weight is an inherent component of the first-order approximation, and omitting it immediately invalidates the token-level optimization objective.

\item Applying R3 in on-policy training does not yield performance gains, despite effectively reducing the training–inference discrepancy (as reflected by the training–inference KL divergence).
Moreover, combining R3 with length normalization even degrades the benchmark score further, and applying R3 without the training-inference IS correction still fails rapidly\footnote{
Our findings regarding R3 under the on-policy setting contradict those reported in \citet{r3}, which may stem from two factors:
(1) They only validate R3 on small-scale experiments, with a maximum of 180 global steps;
(2) Our FP8 inference setting imposes a more stringent stress test on algorithmic correctness, which cannot be adequately captured under their BF16 inference setting.
}.
This empirically confirms our speculation in \S~\ref{subsec:routing_replay}—that Routing Replay can alter the original target policy and introduce bias into the optimization objective.

\end{itemize}

These results demonstrate that, in designing token-level optimization objectives, only those that preserve the validity of the first-order approximation to the expected sequence-level reward lead to improved training stability and performance.
This also validates the soundness of our proposed formulation.

\subsection{Results of Off-policy Training}
\label{subsec:off-policy}

The inference time in RL is typically bounded by the generation length and cannot be accelerated by increasing computational resources.
To leverage increased compute for faster convergence, a common practice is to introduce off-policy updates.
Within a synchronous RL framework, this means that a large batch of responses is split into \(N\) mini-batches for multiple gradient updates.
To investigate the recipes for stable RL training under off-policy settings, we experiment with three levels of off-policiness: with the mini-batch size fixed at 1,024 responses, the global batch size is varied to 2,048, 4,096, and 8,192, corresponding to \(N=2\), \(4\), and \(8\), respectively.
With MiniRL as the baseline, we compare the following methods: MiniRL (no clipping), MiniRL + R2  (no clipping), MiniRL + R2, and MiniRL + R3.

\begin{figure*}[htbp]
    \centering
    \includegraphics[width=\linewidth]{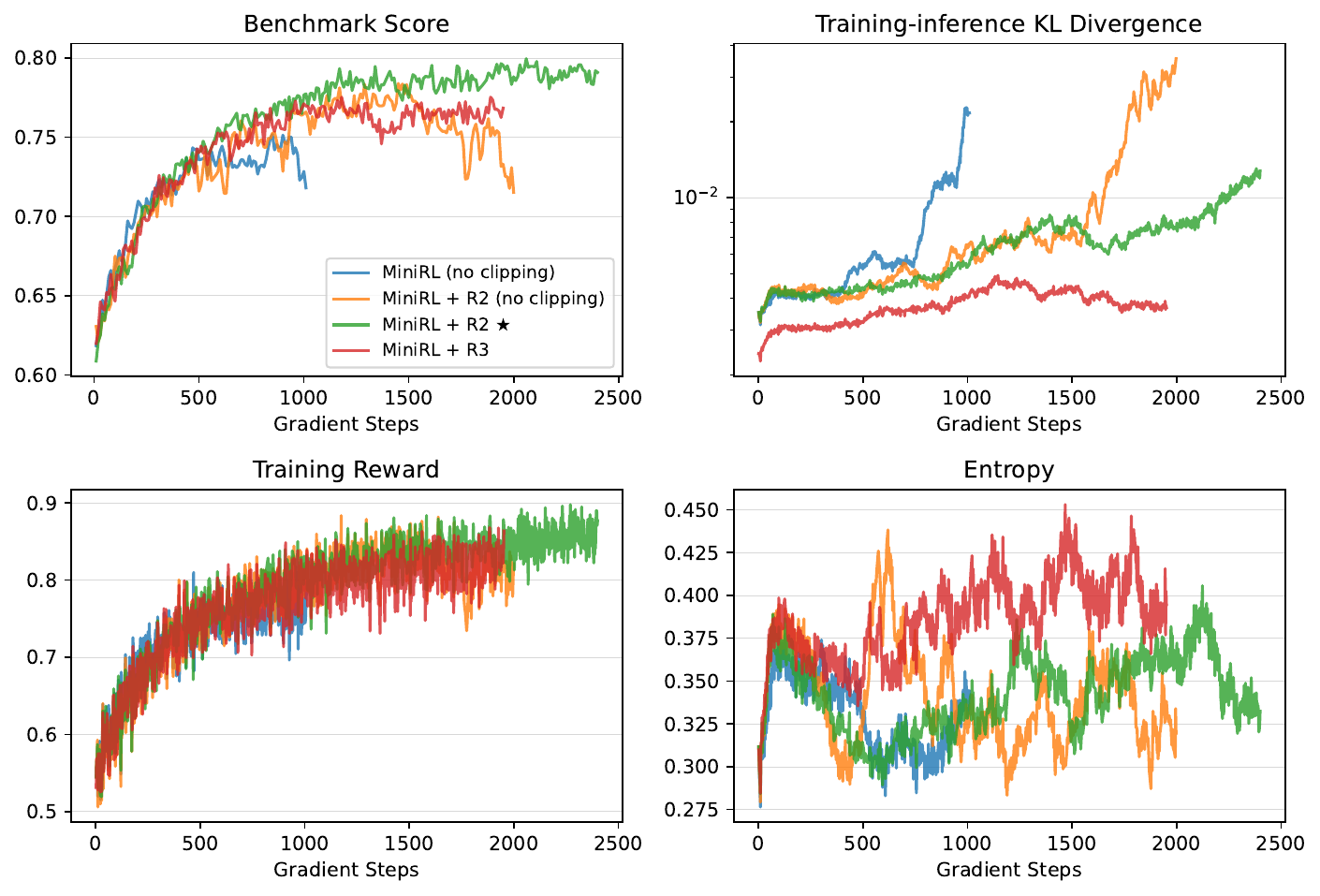}
    \caption{Results of off-policy training with \( \text{gbs}=2\times\text{mbs}=2,048 \).}
    \label{fig:gbs=2mbs}
\end{figure*}

\begin{figure*}[htbp]
    \centering
    \includegraphics[width=\linewidth]{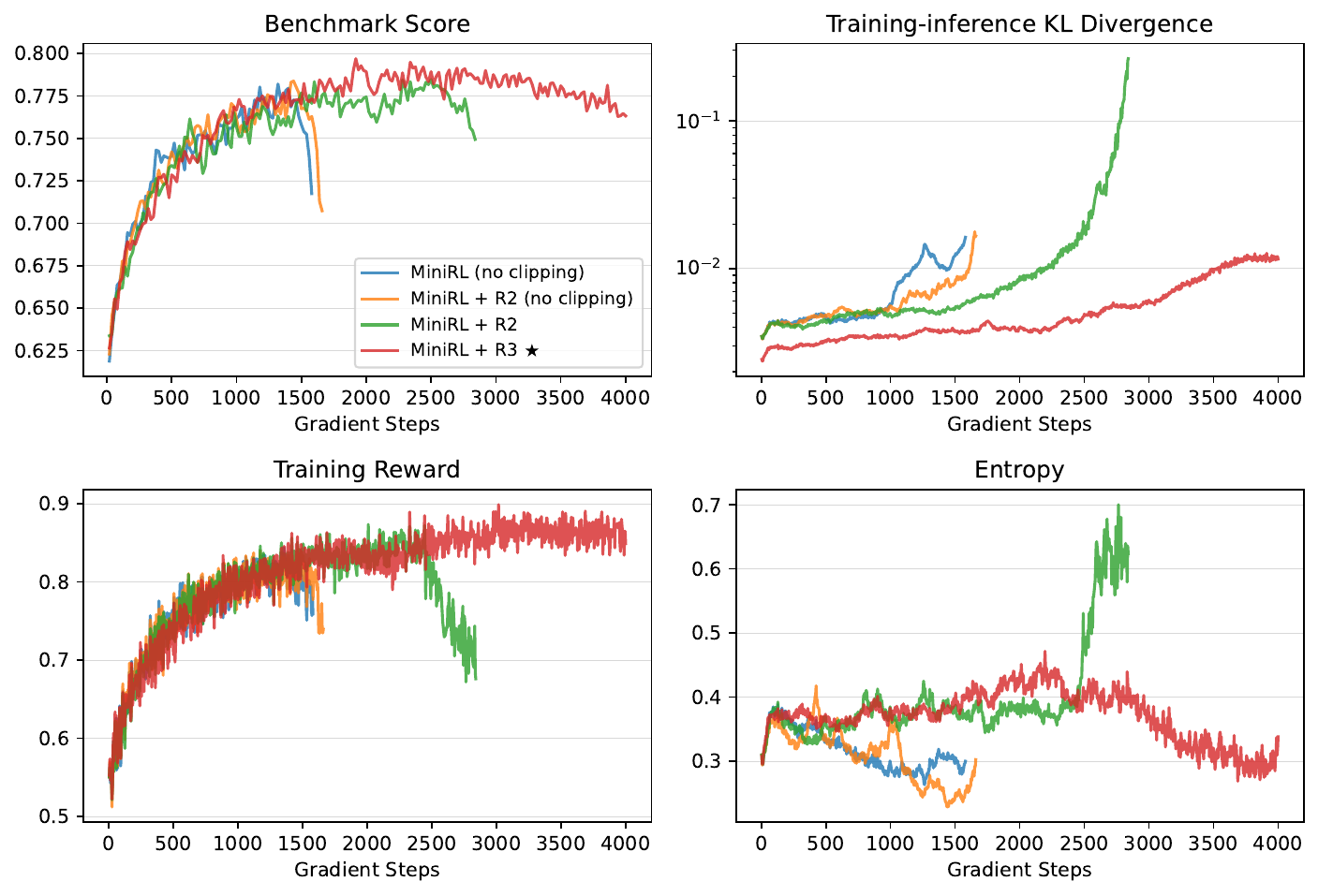}
    \caption{Results of off-policy training with \(\text{gbs}=4\times\text{mbs}=4,096 \).}
    \label{fig:gbs=4mbs}
\end{figure*}

\begin{figure*}[htbp]
    \centering
    \includegraphics[width=\linewidth]{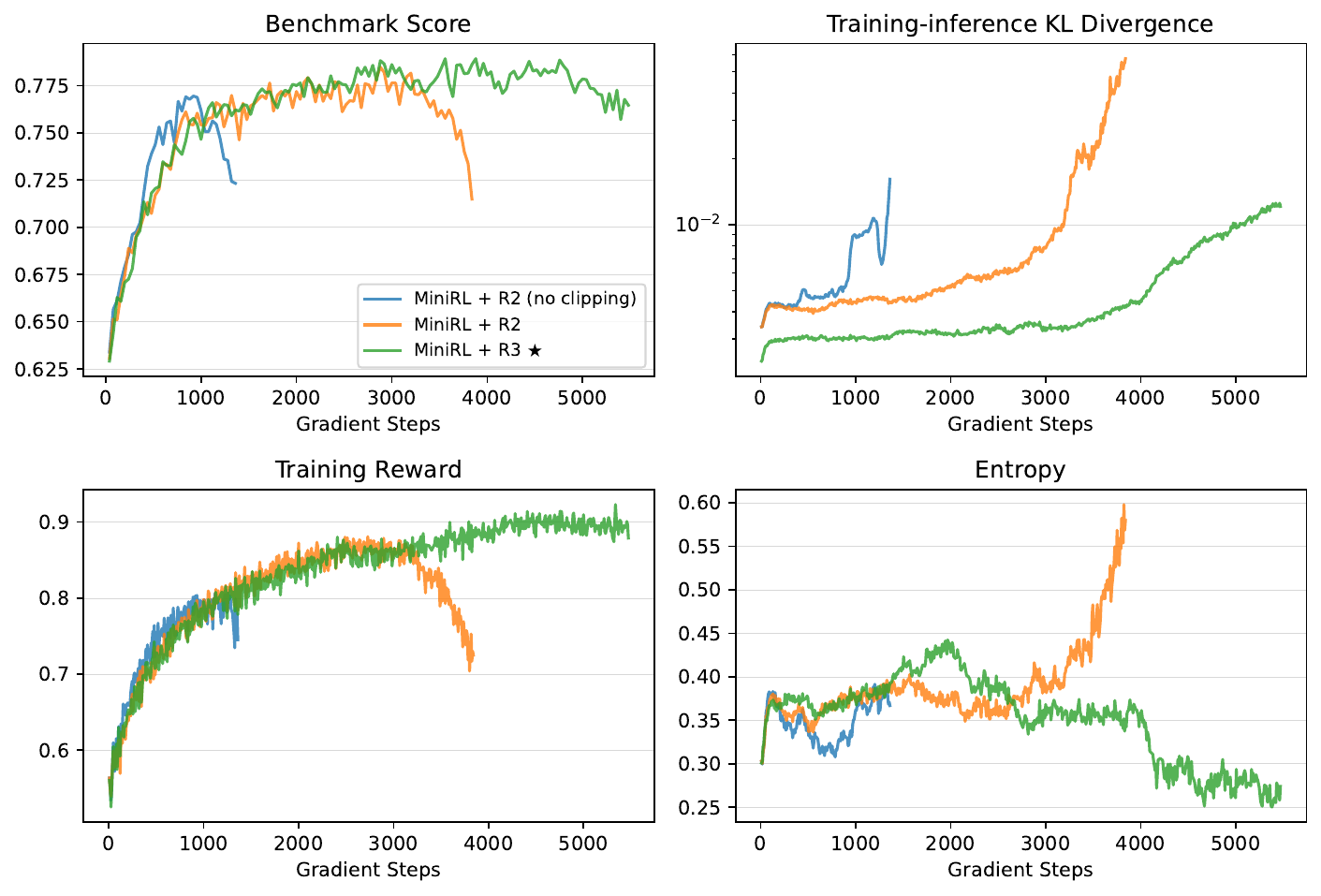}
    \caption{Results of off-policy training with \( \text{gbs}=8\times\text{mbs}=8,192 \).}
    \label{fig:gbs=8mbs}
\end{figure*}

From Figures~\ref{fig:gbs=2mbs} to~\ref{fig:gbs=8mbs}, we draw the following observations and conclusions:
\begin{itemize}[leftmargin=10mm, itemsep=2mm]

\item Once off-policy updates are introduced, both Routing Replay and clipping become essential for stable training.
As shown in Figures~\ref{fig:gbs=2mbs} and~\ref{fig:gbs=4mbs}, omitting either Routing Replay or clipping causes training to collapse prematurely, thereby degrading peak performance.
This indicates that Routing Replay alleviates the impact of expert routing, and the clipping mechanism also effectively prevents aggressive policy updates, thereby both restraining policy staleness.

\item When off-policiness is small (\( \text{gbs}=2\times\text{mbs} \)), R2 outperforms R3, while when off-policiness is large (\( \text{gbs}=4\times\text{mbs} \) and \( \text{gbs}=8\times\text{mbs} \)), R3 surpasses R2.
Notably, under high off-policiness, R2 fails to sustain stable training, and its peak performance achieved before training collapse is also slightly lower than that of R3.
Combining our analysis in \S~\ref{subsec:routing_replay}—particularly that R2 leaves the target policy of the first mini-batch unchanged while R3 alters it—and the on-policy experimental results in \S~\ref{subsec:on-policy}, we hypothesize that when off-policiness is small, the detrimental impact of R3's alteration to the target policy outweighs its benefit in preserving the validity of the first-order approximation, while under larger off-policiness, the opposite holds true.

\end{itemize}

In summary, we find that Routing Replay and clipping are necessary for stable off-policy training.
When off-policiness is small, R2 is sufficient and more effective at stabilizing RL training for MoE models, whereas R3 becomes necessary under larger off-policiness.

\subsection{Results of Varying Cold-start Initializations}

Recall the motivation for stabilizing RL training: given a base model, once we can reach its performance limit through sufficiently long RL training, we can \textit{reliably} enhance the model's capabilities by investing computational resources into RL.
To this end, we investigate whether models initialized with different cold-start data can achieve similar performance when trained using a stable RL recipe.
We compare three versions of cold-start data distilled from three frontier models: Qwen3-Max-Thinking-Preview, DeepSeek-R1-0528, and gpt-oss-120b (high mode).
We report results based on an early-experimental small Qwen3Next MoE model, trained with a global batch size of 4,096, a mini-batch size of 2,048 (\(B=128, G=16, N=2\)), and a generation length of 65,536 tokens.
We employ MiniRL + R2 as the training recipe.

\begin{figure*}[htbp]
    \centering
    \includegraphics[width=\linewidth]{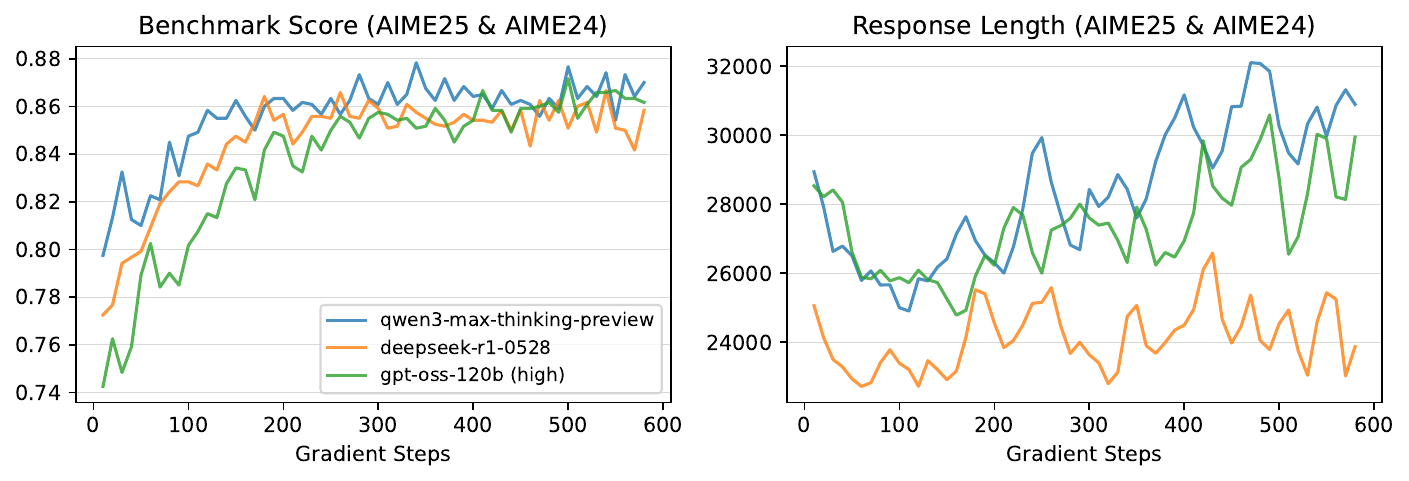}
    \caption{Results of varying cold-start initializations.}
    \label{fig:cold-start}
\end{figure*}

In Figure~\ref{fig:cold-start}, we show that the three cold-start initializations consistently achieve comparable final performance, which encourages us to focus more on RL itself rather than overly on the specifics of cold-start initialization.
Furthermore, comparing Figures~\ref{fig:gbs=mbs} to~\ref{fig:gbs=8mbs}, we find that both on-policy and off-policy training—once stabilized—also consistently achieve similar peak performance.
These results further suggest that stable training plays a decisive role in successfully scaling RL.

\section{Conclusion}

We propose a new formulation for reinforcement learning with LLMs, viewing the token-level optimization objective as a first-order approximation to the true expected sequence-level reward.
Through extensive experiments, we demonstrate that techniques that preserve the validity of this first-order approximation—such as importance sampling correction, clipping, and Routing Replay for MoE models—all effectively stabilize RL training.
We further investigate recipes for stable RL training across varying degrees of off-policiness and show that, once training is stabilized, the same base model consistently converges to similar performance with prolonged RL.
We hope that the insights and empirical results shared in this paper will inspire and facilitate future research.

\bibliographystyle{plainnat}
\bibliography{reference}

\appendix

\section{Comparison of MiniRL against GRPO and CISPO}
\label{app:algo}

We compare the optimization objective of MiniRL against those of GRPO \citep{grpo} and CISPO \citep{minimax-m1}.
With the notations in this paper, GRPO employs the following objective:
\begin{align*}
\mathcal{J}_\text{GRPO}(\theta) 
= &\
\mathbb{E}_{ x \sim \mathcal{D},\, \{y_i\}_{i=1}^G \sim \mu_{\theta_\text{old}}( \cdot | x) } \\
&\ \quad 
\left[ \frac{1}{G} \sum_{i=1}^{G} \frac{1}{|y_i|} \sum_{t=1}^{|y_i|} 
\min \left( r_{i,t}(\theta) \widehat{A}_{i,t},  \, \mathrm{clip} ( r_{i,t}(\theta), 1 - {\varepsilon_\text{low}}, 1 + {\varepsilon_\text{high}} ) \widehat{A}_{i,t} \right)
\right],
\end{align*}
and CISPO is as follows:
\begin{align*}
\mathcal{J}_\text{CISPO}(\theta)
= &\
\mathbb{E}_{ x \sim \mathcal{D}, \, y \sim \mu_{ \theta_\text{old} } ( \cdot | x ) } \\
&\ \quad
\left[ \frac{1}{\sum_{i=1}^{G} |y_i|}
\sum_{i=1}^{G}  \sum_{t=1}^{|y_i|} 
\mathrm{sg} \left[ \mathrm{clip} ( r_{i,t}(\theta), 1 - {\varepsilon_\text{low}}, 1 + {\varepsilon_\text{high}} ) \right] 
\widehat{A}_{i,t} \, 
\log \pi_\theta ( y_t | x, y_{<t} ) \right],
\end{align*}
where in both objectives:
\begin{align*}
r_{i,t}(\theta)=\frac{ \pi_{\theta} (y_{i,t} | x, y_{i,<t}) }{ \pi_{\theta_\text{old}} (y_{i,t} | x,y_{i,<t})},\squad
\widehat{A}_{i,t} = \frac{ R(x, y_i) - \mathrm{mean} \left( \{ R(x, y_i) \}_{i=1}^G \right) }{ \mathrm{std} \left( \{ R(x, y_i) \}_{i=1}^G \right) }.
\end{align*}

Their key differences from MiniRL include the following:
(1) Their original objectives do not consider the training–inference discrepancy;
(2) They both employ length normalization, which we show in \S~\ref{subsec:on-policy} invalidates the first-order approximation to the true expected sequence-level reward and can lead to a biased token-level optimization objective and suboptimal performance;
(3) CISPO does not clip the gradient of certain tokens, which we show in \S~\ref{subsec:off-policy} can result in unstable training.

\section{Detailed Benchmark Results}

\begin{figure*}[htbp]
    \centering
    \includegraphics[width=\linewidth]{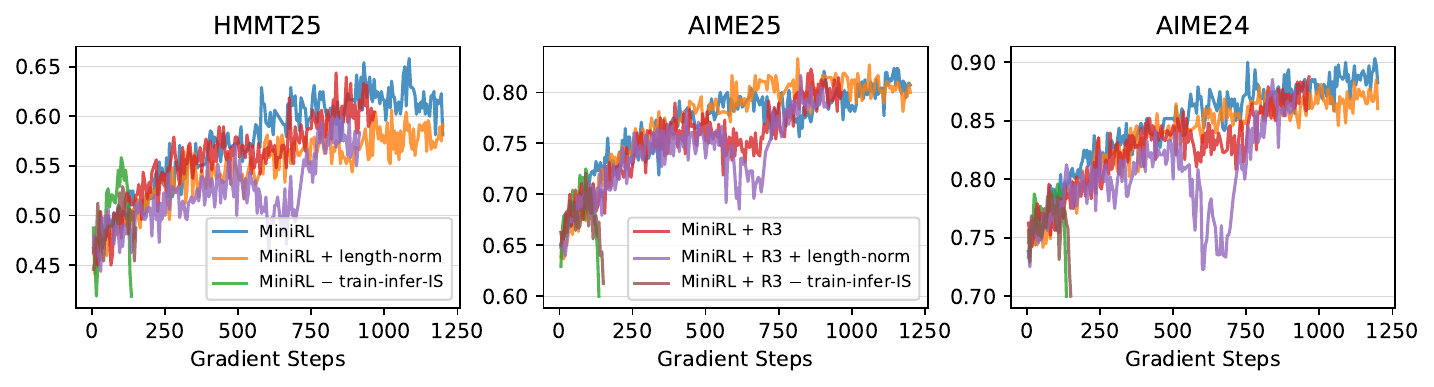}
    \caption{Detailed benchmark results of on-policy training with \( \text{gbs}=\text{mbs}=1,024 \).}
    \label{fig:app_gbs=mbs}
\end{figure*}

\begin{figure*}[htbp]
    \centering
    \includegraphics[width=\linewidth]{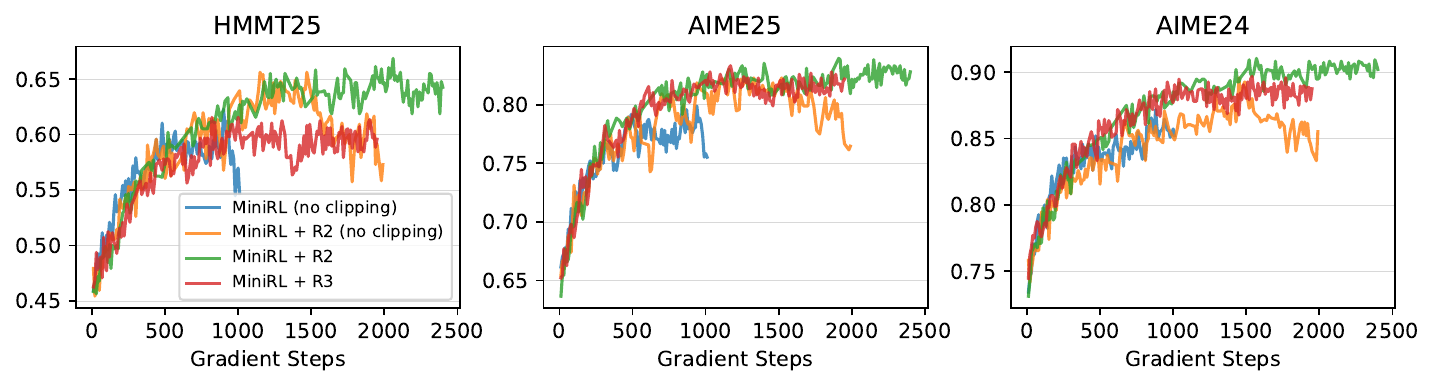}
    \caption{Detailed benchmark results of off-policy training with \( \text{gbs}=2\times\text{mbs}=2,048 \).}
    \label{fig:app_gbs=2mbs}
\end{figure*}

\begin{figure*}[htbp]
    \centering
    \includegraphics[width=\linewidth]{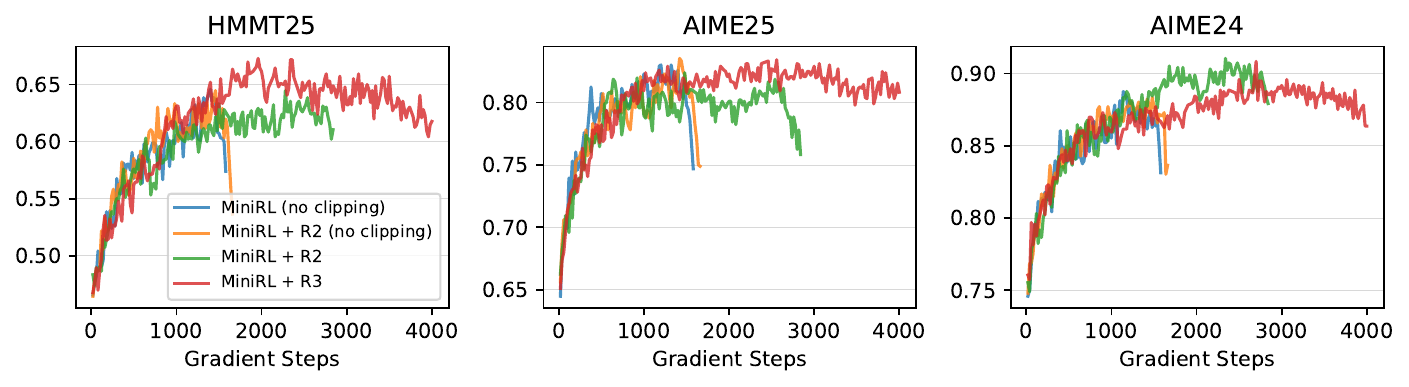}
    \caption{Detailed benchmark results of off-policy training with \(\text{gbs}=4\times\text{mbs}=4,096 \).}
    \label{fig:app_gbs=4mbs}
\end{figure*}

\begin{figure*}[htbp]
    \centering
    \includegraphics[width=\linewidth]{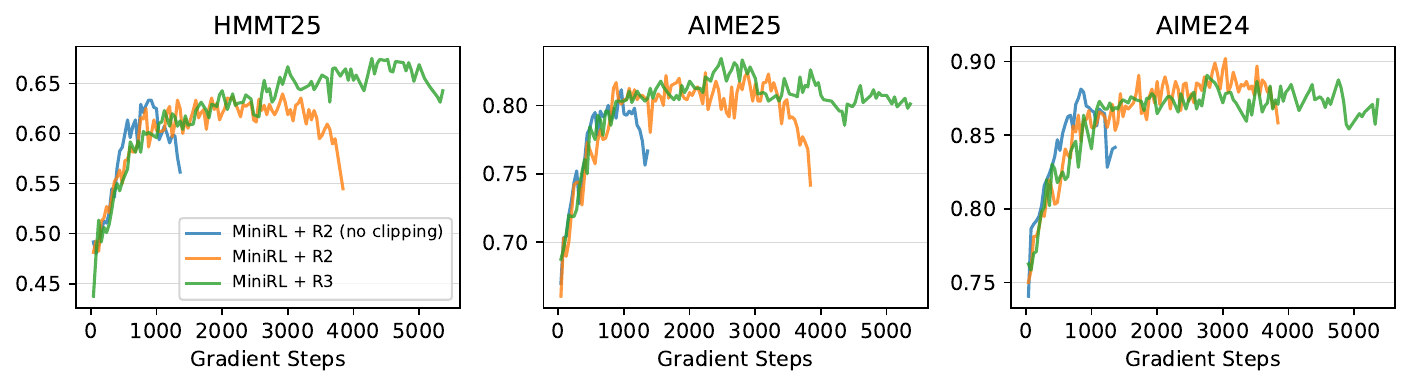}
    \caption{Detailed benchmark results of off-policy training with \( \text{gbs}=8\times\text{mbs}=8,192 \).}
    \label{fig:app_gbs=8mbs}
\end{figure*}

\begin{figure*}[htbp]
    \centering
    \includegraphics[width=\linewidth]{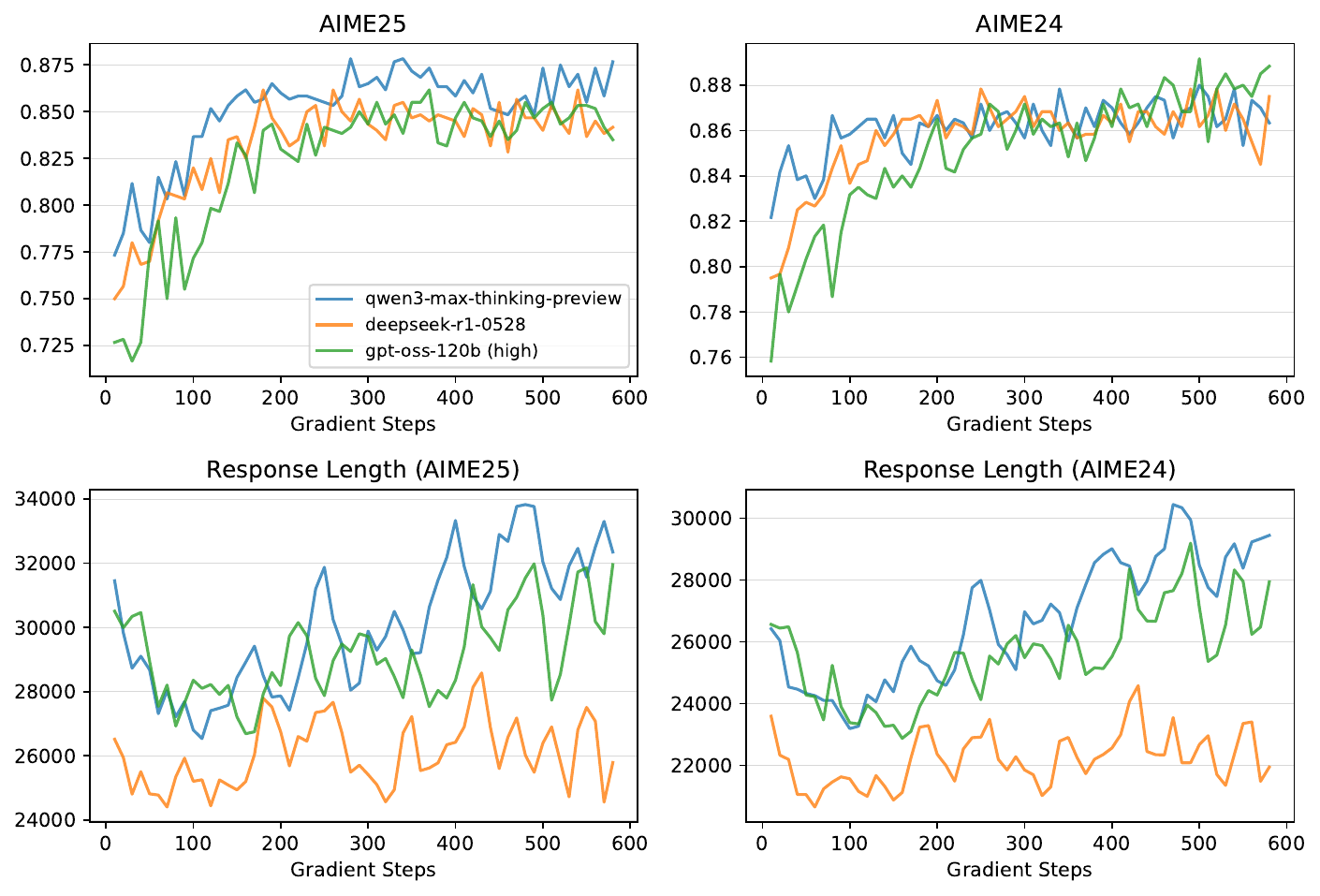}
    \caption{Detailed benchmark results of varying cold-start initializations.}
    \label{fig:app_cold-start}
\end{figure*}

\end{document}